# Australia's long-term electricity demand forecasting using deep neural networks

Homayoun Hamedmoghadam[2*], Nima Joorabloo[1*] and Mahdi Jalili[1#],

*1 School of Engineering, RMIT University, Melbourne, Australia*

*2 Faculty of Engineering, Monash University, Melbourne, Australia*

*\* These two authors contributed equally to this work.*

*# corresponding author: Mahdi.jalili@rmit.edu.au*

**Abstract**

Accurate prediction of long-term electricity demand has a significant role in demand side management and electricity network planning and operation. Demand over-estimation results in over-investment in network assets, driving up the electricity prices, while demand under-estimation may lead to under-investment resulting in unreliable and insecure electricity. In this manuscript, we apply deep neural networks to predict Australia's long-term electricity demand. A stacked autoencoder is used in combination with multilayer perceptrons or cascade-forward multilayer perceptrons to predict the nation-wide electricity consumption rates for 1-24 months ahead of time. The experimental results show that the deep structures have better performance than classical neural networks, especially for 12-month to 24-month prediction horizon.

# 1. Introduction

Power grids are one of the most critical infrastructures and have a major role in sustainable development and economic growth. Smart grids are the future technologies in power grid development, management, and control [1-3]. They have revolutionized the regime of existing power grids, by employing advanced monitoring, communication and control technologies to provide secure and reliable energy supply. Traditional operation of distribution systems in power grids is a top-down oriented and unidirectional, where the operations are set by distribution companies without any kind of engagement from the customers' side. New technologies have changed energy consumption, making it necessary to use effective energy management strategies. For example, in residential areas with many rooftop photovoltaic cells installed, during daytime there might be a reverse power flow from the original power system design [4]. Such a reverse flow can create problems such as voltage and frequency distortions if not properly managed. Another example is electric vehicles with vehicle-to-grid capability that can benefit the demand management by providing certain services for load shifting and powerline congestion management [5]. Integration of information and communication infrastructures with power grids enables considering the load as an additional degree of freedom in network management and makes the grid "smart" by properly engaging the customers in the decision making process [6].

In order to have an effective demand side management strategy and also efficient network planning and operation, reliable short- and long-term load forecasting is essential [7]. Distribution-level short-term load forecasting systems predict the load of substations, feeders and individual customers [8]. Short-term load forecasting is often performed for half an hour to one week ahead time intervals. Individual loads often have abrupt temporal variations due to changes driven by building stock variation and customers' behavioral changes [9-11]. This makes the short-term load forecasting at an individual level a challenging task that many conventional prediction models fail to provide accurate predictions. Often, peaks are overestimated to accommodate generation margins, resulting in non-efficient use of resources [9]. Load patterns at an individual level include high levels of uncertainty and prediction models should properly consider such volatility and uncertainty. A solution to cancel out (or minimize) the uncertainty is to perform the prediction task at an aggregated level, as aggregated loads have more regular patterns than individual loads [8, 12-14].

Forecast terms from one month to one year are considered mid-term and those for the future periods of more than one year are long-term. Both mid- and long-term load forecasts have significant importance for network strategic planning and development of the grid such as scheduling of maintenance activities, installation of new generation and transmission capacities, and long-term demand side management [15, 16]. There are a number of socioeconomic factors influencing the long-term electricity load of the network [17, 18]. For example, while industrialization and rural electrification have a major role in long-term demand increase in developing nations, increased use of electric appliances is one of the major factors behind increased demand in developed nations [16]. Lee and Hong used fuzzy logic and introduced a hybrid mid-term demand forecasting that could successfully predict the electricity demand for periods of up to four-months ahead [19]. Smart grids have many probabilistic factors making the demand forecasting a more challenging problem than before. Liu et al. proposed a probabilistic model for long-term demand forecast and showed it such a framework can provide a reliable prediction [20].

Network planning is a forward-looking process that needs short- and long-term load forecast at the same time. A recent report carried out by the Australian Chief Scientists, known as Finkel Report [21], pinpoints inaccurate demand forecasting as one of the main reasons behind some of Australian electricity grid problems. Demand over-estimation may lead to wasteful and inefficient over-investment in the assets, thus driving up the costs, which is passed onto customers making the energy prices less affordable. On the other hand, demand under-estimation may contribute to inadequate investment that can possibly result in energy reliability and security issues.

This manuscript studies the potential of deep neural networks to predict Australia's long-term electricity demand. Deep learning techniques have been successfully applied to a number of modelling and classification tasks [22-25]. Recently, deep learning has also been applied for short-term demand forecasting. Dedinec et al. used deep belief networks that are made up from multiple layers of Boltzmann machines for short-term demand forecasting [26]. They showed that such a framework can obtain better results than traditional multilayer neural networks. Kong et al. proposed long short-term memory recurrent neural networks for residential short-term demand forecasting [27]. Ryu et al. used deep neural networks for load forecasting and demand side management and showed their better performance than shallow neural networks [28]. Shi et al. used pooling-based deep recurrent neural networks for short-term load forecasting of individual households [29]. Their proposed framework outperformed traditional recurrent neural networks. This manuscript is the first attempt to apply deep neural networks for long-term demand forecast. Our results show that deep neural networks outperform shallow networks for long-term forecasting.

## 2. Methods

### 2.1. Datasets

In this study, our problem is to predict the monthly amounts of country-wide electricity consumption in Australia. We also account for socio-economic and environmental factors. We use publicly available monthly record of electricity demand reported by the Australian Energy Market Operator (AEMO), which is the regulatory body responsible for operating Australia's largest electricity and gas markets and power systems. AEMO's operations include National Electricity Market, the interconnected power grid for Australia's eastern and south-eastern areas, and Wholesale Electricity Market that operates in state of Western Australia.

The socio-economical inputs to our model are Gross Domestic Product (GDP) and population. We use yearly Australia's GDP and population provided as time series data in the World Bank's data bank. Environmental inputs of the model are monthly precipitation, average temperature, average minimum temperature, and average maximum temperature. Precipitation time series is the monthly average value in depth over space within Australia, provided by the World Bank. Land surface temperature data used in the study is reported as anomalies relative to the average of monthly values from January 1951 to December 1980 adopted from Berkeley Earth data collection. We also use the amount of Carbon dioxide emissions during a year in the country as an independent variable to model electricity demand. The data used here covers the time frame from January 1999 to August 2013.

## 2.2. Modeling framework

In order to model the electricity demand as a dependent variable based on a number of independent variables, we utilize deep and shallow artificial neural networks with different architectures and settings. The primary input variables considered in this study are socio-economic and environmental factors, assumed to be relevant and influential to the amount of electricity consumed by residential and industrial. Independent variables, namely GDP, population, $CO_2$ emissions, precipitation, average temperature, average high temperature, and average low temperature are fed to the model as inputs and the model is trained to output the estimated monthly electricity demand. Each sample in the data is associated with a particular month within the interval starting from January 1999 until August 2013 (176 months) which makes the sample size equal to 176. The time series of the dataset can be shown as $(X_i, y_i)$, $i = 1, 2, \ldots, 176$, where $X_i$ is a vector of inputs and $y_i$ is the corresponding output.

A proportion of the available data is used for supervised learning, which is called the training time window here. For each sample within the training time window, i.e. a set of the inputs and the output associated with a particular month, the model is used to estimate the output with the given input values. The error of estimated target value is used to adjust the weights of the links connecting neurons of different layers. The neural network is expected to learn the complex relation between the independent variables (inputs) and the dependent variable (output) through the supervised learning procedure by adjusting its synaptic weights.

Through the experiments we perform a supervised learning process using samples within a certain time window of length $l$, always starting with the first sample (first month of the data); i.e. $(X_i, y_i)$, $i = 1, 2, \ldots, l$. In order to test the trained model based on the training dataset, we feed the neural network with the input vector of samples associated with 1 to 24 months ahead of the last month within the training time window, i.e. $X_i$ with $i = l + 1, l + 2, \ldots, l + 24$. Then, the estimated values by the neural network, i.e. $\hat{y}_i$, $i = l+1, l+2, \ldots, l+24$, are compared against the corresponding target values, i.e. $y_i$, $i = l + 1, l + 2, \ldots, l + 24$, and the estimation error is calculated as a measure of performance (prediction accuracy) for the model.

## 2.3. Evaluation Criteria

In the following experiments, one or more of three widely-used accuracy measures are used, namely, Mean Absolute Error (MAE), Root Mean Squared Error (RMSE), and Mean Absolute Percentage Error (MAPE) defined as below:

$$e_{MAE} = \frac{1}{N}\sum_{i=1}^{N}|y_i - \hat{y}_i| \tag{1}$$

$$e_{MAPE} = \sqrt{\frac{1}{N}\sum_{i=1}^{N}(y_i - \hat{y}_i)^2} \tag{2}$$

$$e_{MAPE} = \frac{100}{N}\sum_{i=1}^{N}\frac{y_i - \hat{y}_i}{y_i} \tag{3}$$

where $y_i$ is the actual value of the output, $\hat{y}_i$ is the forecasted (or estimated) value of the output, and $N$ is the number of test samples. Smaller error values correspond to better prediction accuracy.

## 2.4. Neural network architectures

### 2.4.1. Conventional artificial neural networks

Artificial Neural Networks (ANNs) are powerful computational tools for modeling and estimation that have been applied to many applications [30]. As conventional neural networks, we consider two models: Multi-Layer Perceptron (MLP) and Cascade-Forward Multi-Layer Perceptron (CFMLP). These two models are two well-known neural network architectures, widely used for non-linear modeling. Both MLP and CFMLP consist of an input (output) layer with number of neurons equal to the number of independent (dependent) variables. They can incorporate any number of hidden layers each with one or more neurons, between their input and output layers. MLP and CFMLP are feed-forward architectures meaning that information propagates in only one direction from input layer towards output layer. In other words, there are no cycles in their networked structures. In MLP neurons in each layer are connected only to neurons of the immediate successive layer (see Fig. 1.a), whereas in CFMLPs neurons in each layer are connected to neurons of all the layers to the front (Fig. 1.b), i.e. towards the output layer.
ANNs have often many parameters that should be optimized based on the dataset in hand. The free parameters of the model are learnt such as a loss function (e.g. prediction error) is minimized. In order to enhance generalizability of the model, often regularization terms are also considered in the optimization process. A number of optimization approaches, with different computational and memory requirements, have been developed for this purpose. Here, we apply a number of optimization (or training) algorithms and choose the best performing one. Using 13 training time window sizes, where smallest window can be described with $l = 120$ and the largest window with $l = 132$, different training algorithms are applied to adjust the networks' weights and biases. Trained models are then used to predict the electricity demand for 24 months ahead of their corresponding training time window, and the average prediction accuracy is used to compare the effectiveness of different training algorithms.

The training algorithms we use are [30]: (*i*) Simple loss function of squared errors and Levenberg-Marquardt optimization algorithm denoted as LM, (*ii*) LM with Bayesian regularization term aiming at better generalization of the model denoted as LMbr, (*iii*) simple loss function and Gradient Descent optimization algorithm denoted as GD, (*iv*) GD using momentum (GDm) which allows the algorithm to escape the shallow local minima, (*v*) GD with adaptive learning rate (GDa) which balances the stability of the solution and fast convergence of the algorithm by changing the learning rate during the training process, (*vi*) GD with momentum and adaptive learning rate denoted as GDma, (*vii*) Conjugate Gradient algorithm which periodically switches the search direction with Powell-Beale reset rule denoted as CGpb, (*viii*) conjugate gradient algorithm using Fletcher-Reeves update rule to determine the new search direction denoted as CGfr, (*ix*) conjugate gradient with Polak-Ribiére updates denoted as CGpr, (*x*) scaled conjugate gradient denoted as SCG, (*xi*) BFGS quasi-Newton method which is an alternative to conjugate gradient methods with more computational complexity but converges faster and here denoted as BFGS, (*xii*) one-step secant optimization method which is improved BFGS algorithm to have less computational and storage requirements here denoted as OSS, and (*xiii*) resilient backpropagation algorithm (RBP) aiming at solving the problem of small magnitude gradient which leads to slow convergence.

### 2.4.1. Deep neural networks

Deep Neural Networks (DNNs) are a particular type of ANNs with multiple layers which apply non-linear operations to data in order to automatically represent the complicated functions via high-level abstraction. One approach to create deep architectures is to use Autoencoders, also called Autoassociators, and wire them with conventional classifiers or regression models. Autoencoders are a class of neural networks with the ability to encode the input data into a new representation with unsupervised learning [25, 31]. An autoencoder is a hidden layer of neurons trained to encode the raw input data into a new representation and decode them back to reconstruct the original input with the minimum deformation possible (Fig. 2). A stacked autoencoder is trained layer-wise in an unsupervised manner; each time a layer is trained to reconstruct its inputs through unsupervised learning using samples of raw inputs or the transformed raw inputs by previous layer(s). Then the front building block, which is a classical neural network here, is trained with the features extracted by the last autoencoder as its inputs and the monthly electricity consumption $y_i$ as its target output; this procedure is called pre-training. Pre-training can be followed by adjusting weights in the deep network utilizing backpropagation. This second training phase is called fine-tuning for the deep neural network.

Here we use MLP and CFMLP and stack them with autoencoders to create DNN architectures (Fig. 3), and evaluate them as models for electricity demand prediction. We use different shallow architectures, namely, one-layer MLP, two-layer MLP, one-layer CFMLP, and two-layer CFMLP each with the best performer settings found (i.e. the number of neurons in the layers and the best training algorithm) and stack them with one or two layers of autoencoders. We vary the number of neurons in each layer of autoencoders between 1 and 15. For each one of the four shallow architectures and with one or two layers of stacked autoencoders, the experiments are performed using training time windows with $l = 120$ to $l = 132$ and evaluation is done according to average prediction accuracy calculated over the whole set of tests, i.e. 24 subsequent months of each training time window in 10 independent runs. Also, experiments are repeated once with only a pre-training phase for blocks of deep network, and once following with a fine-tuning phase for the whole deep neural network after pre-training of the blocks using the most efficient backpropagation algorithm found for the classical neural network stacked in the front of the deep structure.

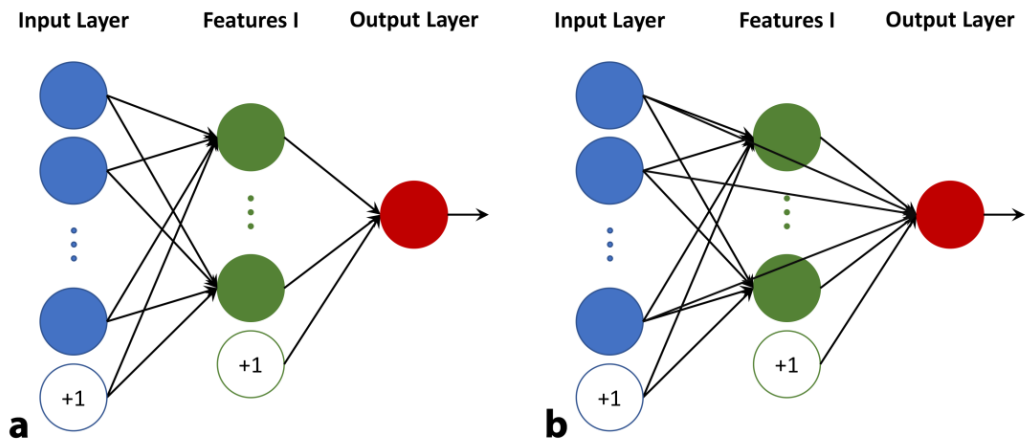

**Fig. 1.** Schematic view of two classical neural network architectures both with a single hidden layer. a) Multi-Layer Perceptron (MLP) and b) Cascade-forward multi-layer perceptron (CFMLP) are two widely used feed forward neural network architectures.

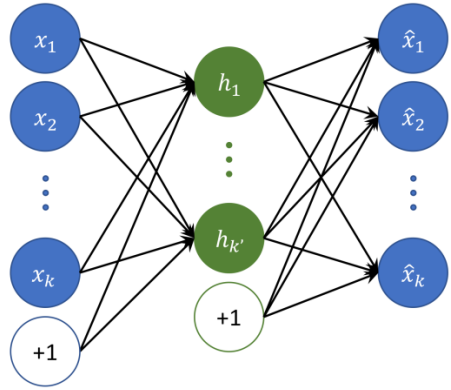

**Fig. 2.** Schematic view of a single autoencoder layer with unsupervised learning process. The autoencoder is fed with a vector of *k* inputs and the network weights are adjusted to encode the inputs to create a new representation with a vector of *k'* values, extracting *k'* features. The autoencoder then decodes the new representation to estimate the inputs with minimum error in the output layer.

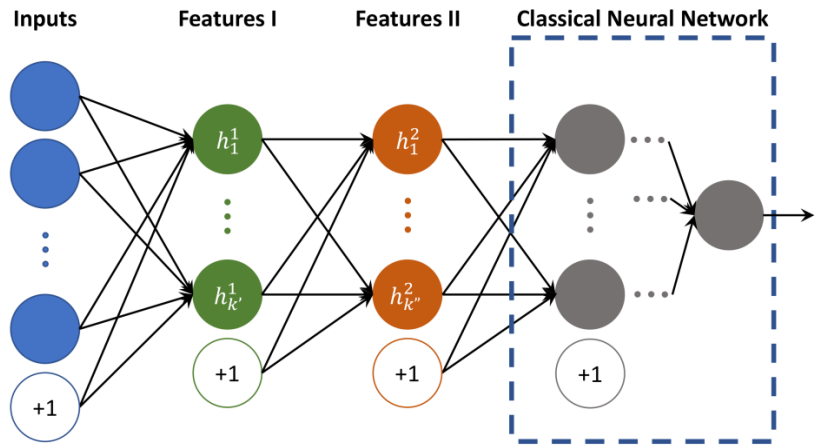

**Fig. 3.** Schematic view of a deep neural network with multiple layers of autoencoders stacked with a classical neural network, e.g. MLP. Autoencoders extract features from the raw inputs and feed the regression model with the new representation of data.

## 3. Results

### 3.1. Conventional artificial neural networks

The first stage in learning an ANN is to find the optimal number of hidden layers and the number of neurons in each layer. In this work, we use only 1 or 2 hidden layers for MLP and CFMLP. In order to tune the model, we perform the learning on MLP and CFMLP architectures with 1 and 2 hidden layers and with different number of neurons in each layer. We then evaluate the model's performance through a set of tests to discover the best performing setting for each architecture. Using the training time window with $l = 120$ (i.e. the last sample as $(X_{120}, y_{120})$), first, the learning process is performed on the models, and then the model with the optimized set of parameters is used to predict the monthly electricity demand for the next consecutive 24 months of the data. For each set of experiments, the training time window is first expanded for a number of times, each time with one month, and then the prediction is performed again for 1 to 24 months ahead of the end of the training time window. Furthermore, for each time window, 10 independent runs of training-then-prediction are performed to make the results more reliable. The prediction accuracy measures are calculated by averaging over all test errors calculated during a set of experiments. A set of experiments includes a range of window sizes with 10 independent runs for each window size.

In order to find the best structure for the two classical ANN architectures, the learning time window is expanded 13 times, so the first and last training time windows have $l = 120$ and $l = 132$, respectively. The learning time window expansion is performed each time by adding a single sample corresponding to the first month. First, a set of experiments are performed using MLP and CFMLP architectures with a single hidden layer. Prediction performance is compared for different number of neurons in the hidden layer to find the best setting for the case of having a single hidden layer (Fig. 4). The results show that when there is only one hidden layer, the prediction performance worsens or no significant improvement is obtained by increasing the number of neurons more than 5. Furthermore, MLP outperforms CFMLP by showing much less prediction error.

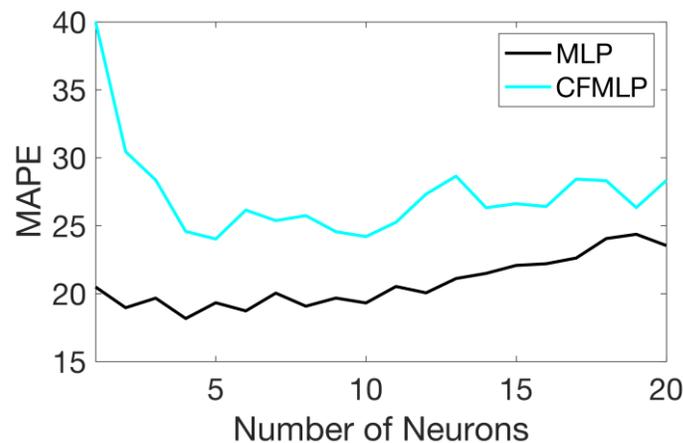

**Fig. 4.** Average prediction error of MLP and CFMLP as classical artificial neural network architectures. The plots show the performance of MLP and CFMLP both having a single hidden layer with different number of neurons. The performance criterion is MAPE calculated for predicting the electricity demand of 1 to 24 months ahead of the learning time window. Neural networks are trained using Levenberg-Marquardt optimization algorithm.

We also performed the experiments on the same learning time window sizes to assess the classical architectures having two hidden layers with the number of neurons varying between 1 and 15 within each hidden layer. According to the results, MLP with more than two neurons in the first hidden layer have relatively good prediction accuracy, and further increase in the number of neurons in the first layer does not affect the performance significantly. However, increasing the number of neurons in the first hidden layer of CFMLP reduces the prediction accuracy. For both MLP and CFMLP architectures, performance usually declines as the number of neurons in the second layer increases more than 2 for MLP and 9 CFMLP. According to the results, the best performance for MLP and CFMLP with single hidden layer is achieved with 4 and 5 neurons respectively. When two hidden layers are used, the best architecture is (5, 2), i.e. 5 neurons in the first layer and 2 neurons in the second layer, for MLP and (2, 9) for CFMLP.
Utilizing different training approaches led to a wide range of prediction performances, according to our experiments (Fig. 5). The results show that effectiveness of different training algorithms is not the same for MLP and CFMLP. The MLP architecture with one hidden layer of 4 neurons achieves the best prediction accuracy when trained with Levenberg-Marquardt algorithm equipped with Bayesian regularization, while resilient backpropagation algorithm is the best learning algorithm for MLP with two hidden layers. The best learning algorithm for CFMLP with one (two) hidden layers is Levenberg-Marquardt algorithm with simple loss function (conjugate gradient with Polak-Ribiére updates). These optimal learning algorithms are used for DNN architectures.

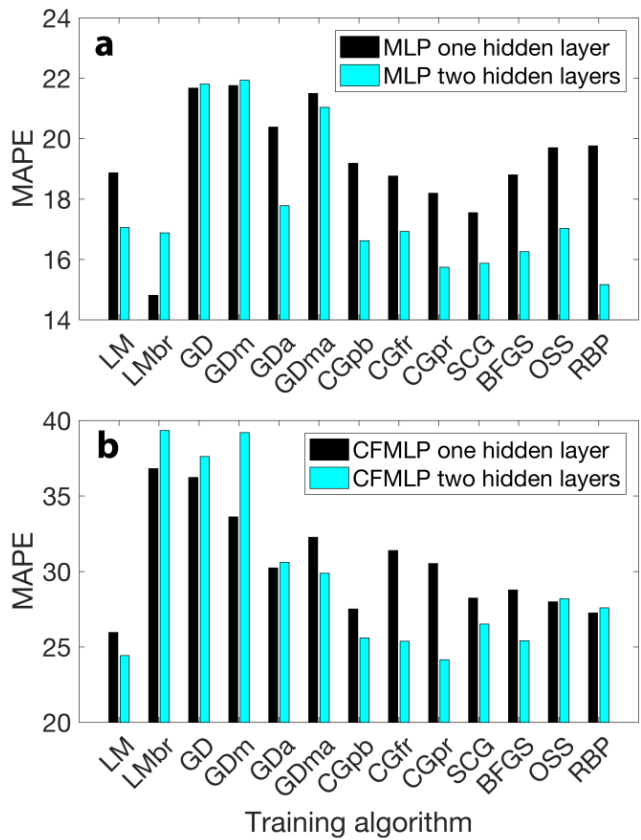

**Fig. 5.** Average prediction error for (a) MLP and (b) CFMLP, achieved by applying different training algorithms. For each architecture, the best-performing structure found through the previous set of experiments is used, i.e. single hidden layer with 4 and 5 neurons for MLP and CFMLP, respectively and (5, 2) neurons and (2, 9) neurons for two-layer MLP and CFMLP, respectively.

### 3.2. Deep neural networks

In this section, we apply the proposed stacked autoencoder with MLP or CFMLP to the data. For MLP with one layer, the best performance is achieved with one layer of stacked autoencoder with 2 neurons. When there are two hidden layers in MLP, the best found structure consists of one layer of stacked autoencoder with 8 neurons. DNNs using the CFMLP with one and two layers have the best prediction accuracy with a single stacked autoencoder layer with 10 and 4 neurons, respectively. All architectures result in higher prediction accuracy when they go through the fine-tuning phase.

We compare the four classical architectures (MLP and CFMLP with one and two hidden layers) with their corresponding deep architectures. Note that the deep networks are built by adding a block of one-layer stacked autoencoder to the classical models with the best-performing settings (i.e. the number of neurons in the hidden layer and the optimal learning algorithm) determined by experimental results. The learning in stacked autoencoder is unsupervised, while supervised learning is used to tune parameters of MLP and CFMLP. In order to compare the classical model with deep networks, we evaluate their prediction error when the demand is forecasted 1-24 months ahead.

Figure 6 shows the prediction error as a function of the length of look-ahead time. It is expected that as prediction horizon is extended, the performance worsens, i.e. the prediction error increases. The rate of decline in performance by increasing the prediction horizon is significantly less in deep networks as compared to their classical counterparts. The results suggest that deep neural networks are always more effective for long-term prediction, i.e. more than 1 year to 2 years ahead, rather than the prediction for up to 1 year ahead (mid-term prediction). Deep networks with one layer of MLP or CFMLP have always better prediction performance than classical MLP or CFMLP, where the outperformance of deep CFMLP over its classical counterpart is more significant than that of MLP. Deep MLP with two hidden layers and deep CFMLP with one layer show the best overall performance, while the deep MLP with two layers is almost robust against increasing the prediction horizon. Overall prediction error is also calculated for different structures by averaging over the prediction horizons (i.e. 24 values each to corresponding to different prediction horizon), and the results are shown in Table 1. It indicates that deep CFMLP with one hidden layer has the best performance in terms of all evaluation criteria (MAPE, RMSE and MAE). These results also show that stacking a layer of autoencoder almost always enhances the overall performance of a classical neural network.

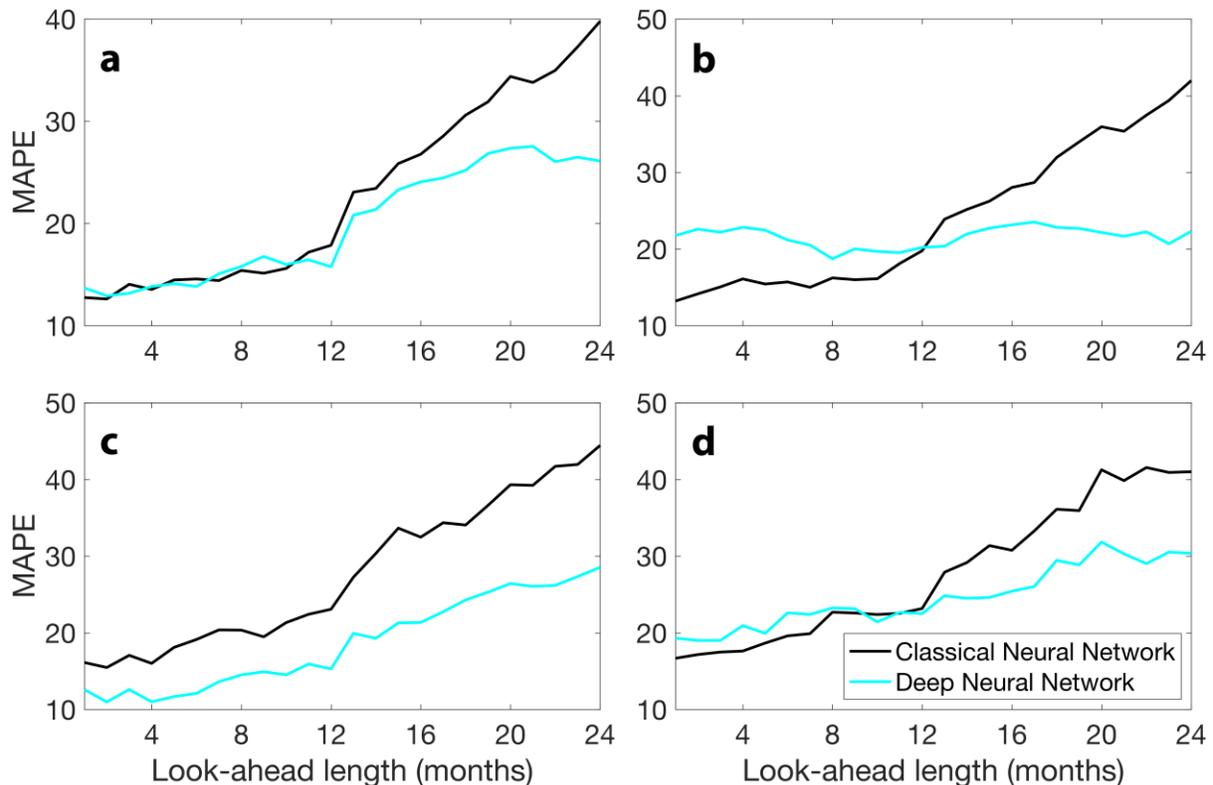

**Fig. 6.** Classical neural networks (MLP and CFMLP) and their corresponding deep architectures. Deep network structures are built by stacking an autoencoder to extract features from the raw inputs and feed the classical neural network with the new representation. The plots show the average prediction error for 1 to 24 months ahead. The classical networks are a) MLP with one hidden layer, b) MLP with two hidden layers, c) CFMLP with one hidden layer, and d) CFMLP with two hidden layers.

**Table 1.** Comparison between classical and deep neural networks in terms of three different evaluation criteria (MAPE, RMSE and MAE) calculated by averaging over all experiments for training time windows with *l* = 133, 134, …, 152 and 10 independent runs for each window size. 24 tests (one for each prediction horizon 1-24 months) are performed for each run of each training window size; an average over these 24 tests is obtained for each run. Error value showing the best performance is shown in **bold**.

|  | MLP with one hidden layer | | MLP with two hidden layers | | CFMLP with one hidden layer | | CFMLP two hidden layers | |
| --- | --- | --- | --- | --- | --- | --- | --- | --- |
|  | Classical | Deep | Classical | Deep | Classical | Deep | Classical | Deep |
| MAPE | 22.83 | 19.88 | 24.13 | 21.59 | 27.68 | **18.68** | 27.90 | 24.66 |
| RMSE | 0.166 | 0.145 | 0.177 | 0.192 | 0.213 | **0.138** | 0.209 | 0.182 |
| MAE | 0.142 | 0.123 | 0.149 | 0.149 | 0.170 | **0.115** | 0.172 | 0.150 |

## 4. Discussion and Conclusion

In this study, we aimed to predict the monthly electricity demand in Australia based on time series of consumption rates as well as socio-economic and environmental factors. We approached the modeling of electricity demand by employing artificial neural networks and predicted the nation-wide electricity consumption for a range from one month (mid-term forecast) to two years (long-term forecast) ahead. Such a long-term demand forecast has significant role in effective electricity network planning and operation. Inaccurate demand forecast has been identified as one of the major reasons behind recent problems with Australian energy systems. This has contributed to reliability and security issues in the energy system, and thus driving up the energy prices.

We collected the electricity consumption, socioeconomic and environmental data from various resources and preprocessed them to make them ready for further processing and neural network modeling. We then carried out a set of tests to find the best performing settings for two widely-used classical neural network architectures, namely, multi-layer perceptron and cascade-forward multi-layer perceptron. Then, both tuned classical architectures with their best performing structures and training approaches, were used with stacked autoencoders as building blocks to create deep neural architectures. Such a deep neural network structure has been successfully used in a number of modeling and classification applications.

We investigated the effect of the number of stacked autoencoder layers with the tuned classical network architectures and found the optimal design settings for deep neural networks. As a final step, the classical neural networks were compared with the corresponding deep architectures built upon them. The comparison was carried out based on the prediction error of the models to predict the electricity demand in a range from 1 to 24 months ahead of the training time window. Experiments revealed the effectiveness of deep neural networks architectures for monthly demand prediction. Stacked autoencoders, when trained in an unsupervised manner to reproduce inputs through an encoding-decoding process, learns to produce an alternative representation of input data. A single layer autoencoder is proved to have the tendency to learn first-order features from the raw input data. Stacked autoencoder feeds the classical neural network block with the new representation of inputs. Our investigations in this manuscript showed how using stacked autoencoders can improve the performance of classical neural network models by capturing latent useful features from the raw inputs. As a future work, similar deep neural network structures will be applied to predict short-term electricity demand in varying granularity levels, from individual household to substations, state and nation-wide levels.

## Acknowledgments
Mahdi Jalili is supported by Australian Research Council through project No DP170102303. The authors thank Dr. Peter Sokolowski (RMIT University, Melbourne, Australia) and Mr. Peter McTaggart (Powercor, Melbourne, Australia) for stimulating discussions on the topic.